\documentclass[letterpaper, 10 pt, conference]{ieeeconf}
\IEEEoverridecommandlockouts
\usepackage[utf8]{inputenc}
\usepackage{graphicx}
\usepackage{multicol}
\usepackage{footmisc}
\usepackage{amssymb}
\usepackage{amsmath}
\usepackage{amsfonts}
\usepackage{algorithm}
\usepackage[noend]{algorithmic}
\usepackage[dvipsnames]{xcolor} 
\usepackage{tikz}
\usepackage{float}
\usepackage{hyperref}
\let\labelindent\relax 
\usepackage{enumitem} 
\usepackage{booktabs} 

\usetikzlibrary{math,shapes,positioning}
\usepackage[font={small}]{caption}
\usepackage[caption=false, font=footnotesize]{subfig}
\usepackage{listings}

\newcommand{\argmin}{\operatornamewithlimits{argmin}}
\newcommand{\irange}[2]{[#1 \nobreak\mathinner{\ldotp\ldotp}\nobreak #2]}
\renewcommand{\vec}[1]{\mathbf{#1}}

\title{\LARGE \bf Robots of the Lost Arc:\\ Self-Supervised Learning to Dynamically Manipulate\\ Fixed-Endpoint Cables}
\author{Harry Zhang, Jeffrey Ichnowski, Daniel Seita, Jonathan Wang, Huang Huang, and Ken Goldberg$^1$%
\thanks{$^1$All authors are affiliated with the AUTOLab at UC Berkeley (automation.berkeley.edu).
{\tt\small \{harryhzhang, jeffi, seita, jnwang19, hh19971229, goldberg\} @berkeley.edu}}}

\begin{document}

\maketitle
\thispagestyle{empty}
\pagestyle{empty}

\begin{abstract}
We explore how high-speed robot arm motions can dynamically manipulate cables to vault over obstacles, knock objects from pedestals, and weave between obstacles. In this paper, we propose a self-supervised learning framework that enables a UR5 robot to perform these three tasks.
The framework finds a 3D apex point for the robot arm, which, together with a task-specific trajectory function, defines an arcing motion that dynamically manipulates the cable to perform tasks with varying obstacle and target locations. 
The trajectory function computes minimum-jerk motions that are constrained to remain within joint limits and to travel through the 3D apex point by repeatedly solving quadratic programs to find the shortest and fastest feasible motion.
We experiment with 5 physical cables with different thickness and mass and compare performance against two baselines in which a human chooses the apex point. Results suggest that a baseline with a fixed apex across the three tasks achieves respective success rates of 51.7\,\%, 36.7\,\%, and 15.0\,\%, and a baseline with human-specified, task-specific apex points achieves 66.7\,\%, 56.7\,\%, and 15.0\,\% success rate respectively, while the robot using the learned apex point can achieve success rates of 81.7\,\% in vaulting, 65.0\,\% in knocking, and 60.0\,\% in weaving.  Code, data, and supplementary materials are available at \url{https://sites.google.com/berkeley.edu/dynrope/home}.
\end{abstract}

\section{Introduction}

Dynamic manipulation and management of linear deformable objects such as ropes, chains, vacuum cords, charger cables, power cables, tethers, and dog leashes are common in daily life. For example, a person vacuuming may find that the vacuum power cable is stuck on a chair, and use dynamic manipulation to ``vault'' the cable over the chair. If the first motion does not succeed, the human can try again, adapting their motion to the failure.  In this paper, we explore how a robot can teach itself to perform three tasks:

\begin{description}[align=left,leftmargin=0pt]
    \item[Task 1: Vaulting] The robot dynamically manipulates a cable to move from one side of an obstacle to another. 
    \item[Task 2: Knocking] The robot dynamically manipulates a cable to knock a target object off an obstacle. 
    \item[Task 3: Weaving] The robot dynamically manipulates a cable to weave it between three obstacles. 
\end{description} 

To accomplish these tasks, we propose to generate trajectories using a parameterized function. To reduce the complexity of parameterizing actions, we compute minimum-jerk trajectories using DJ-GOMP~\cite{GOMP, DJGOMP} that take as input the apex point in the trajectory.  This point, combined with task-specific starting and ending arm configurations, are used to generate a single high-speed trajectory for a physical UR5 robot. 
\begin{figure}
    \centering
    \includegraphics[width=\linewidth]{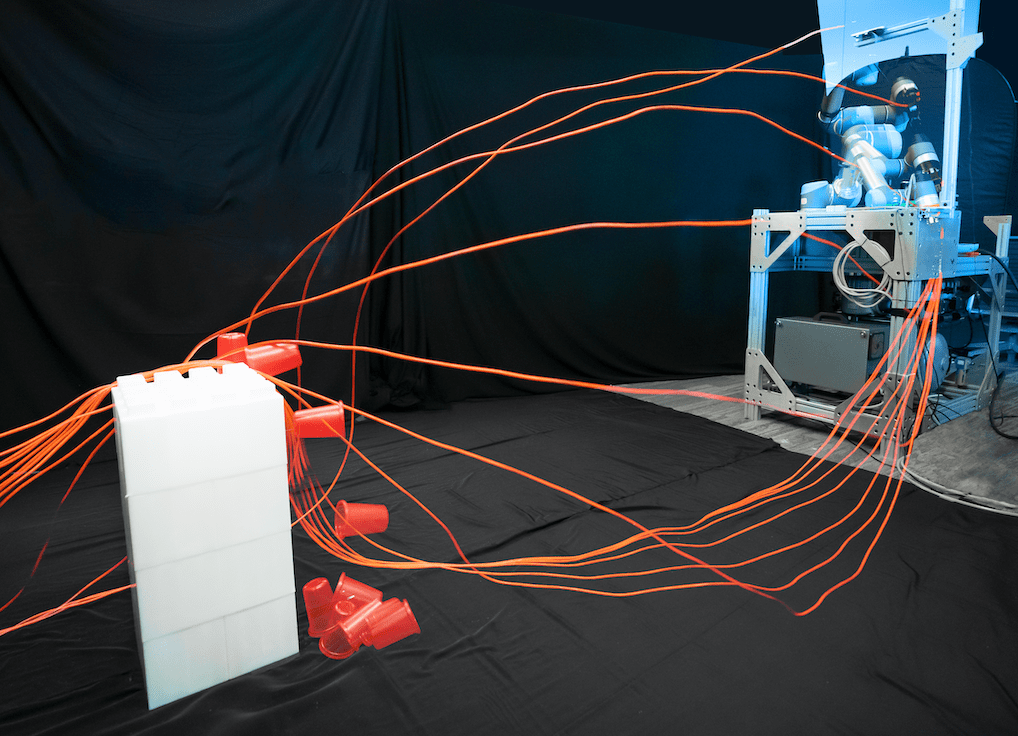}
    \caption{Long exposure photo of a UR5 robot dynamically manipulating an orange cable fixed at one end (far left) to knock the red target cup off the white obstacle using the learned 3d apex point and minimum-jerk robot trajectory.}
    \label{fig:teaser_new}
    \vspace*{-20pt}
\end{figure}
We present a self-supervised deep imitation learning pipeline that learns to predict the apex point given target positions. Using a simulator we develop for vaulting, with a simulated UR5, we obtain an example action that completes the vaulting task. For any of the three tasks in the real world (see Fig.~\ref{fig:teaser_new}), we then have the physical UR5 continually execute actions based on the action computed in simulation, with noise added to increase dataset diversity. While the robot performs actions, one end of the cable is attached to its gripper, and the other is fixed to a wall. (Hereafter, we use \emph{cable} to refer to any 1D deformable object with low stiffness, such as ropes, cords, and strings.) This process results in a self-supervised procedure for collecting a large dataset. We use the resulting data to train a deep neural network policy to predict an action (i.e., the apex point) given a target position. 

This paper contributes:
\begin{enumerate}
    \item Three novel dynamic cable manipulation tasks: vaulting, knocking, and weaving.
    \item INDy, a self-supervised framework for collecting data and learning a dynamic manipulation policy using parameterized minimum-jerk trajectories.
    \item A simulation platform to experiment with novel tasks and policy learning.
    \item Physical experiments evaluating the policies with a UR5 robot on 5 different cables.
\end{enumerate}


\section{Related Work}

Manipulation of 1D deformable objects has applications in surgical suturing~\cite{schulman_case_study_2013,suturing_autolab_2016}, knot-tying~\cite{van_den_berg_2010,case_study_knots_1991,knot_planning_2003,tying_precisely_2016}, untangling ropes~\cite{rope_untangling_2013,grannen2020untangling,tactile_cable_2020}, deforming wires~\cite{wire_insertion_1996,wire_insertion_1997}, and inserting cables into obstacles~\cite{tight_tolerance_insertion_2015}; Sanchez et al.~\cite{survey_2018} provide a survey.

Various approaches have been proposed for quasistatic manipulation of cables, such as using demonstrations~\cite{schulman_isrr_2013,lee_LFD_IROS_2014,seita_bags_2021}, self-supervised learning~\cite{nair_rope_2017,ZSVI_2018, jin2024multi}, model-free reinforcement learning~\cite{lerrel_2020,corl2020softgym, elmquist2022art}, contrastive learning~\cite{yan2020learning, sim2019personalization}, dense object descriptors~\cite{priya_2020, eisner2022flowbot3d, pan2023tax}, generative modeling~\cite{wang_visual_planning_2019, avigal20206, avigal2021avplug}, and state estimation~\cite{state_estimation_LDO, zhang2020dex, zhang2021robots, zhang2023flowbot++, yao2023apla}. These prior works generally assume quasistatic dynamics to enable a series of small deformations, often parameterized with pick-and-place actions. Alternative approaches involve sliding cables~\cite{one_hand_knotting_tactile_2007,zhu_sliding_cables_2019} or assuming non-quasistatic systems that enable tactile feedback~\cite{tactile_cable_2020, jin2024multi}. This paper focuses on dynamic actions to manipulate cables from one configuration to another, traversing over obstacles and potentially knocking over objects.

More closely related is the work of Yamakawa et al.~\cite{dynamic_knotting_2010,yamakawa2012simple,high_speed_knotting_2013,one_hand_knotting_tactile_2007, zhang2016health, devgon2020orienting, lim2021planar, lim2022real2sim2real} that focuses on dynamic manipulation of cables using robot arms moving at high speeds, enabling them to model rope deformation algebraically under the assumption that the effect of gravity is negligible. They show impressive results with a single multi-fingered robot arm that ties knots by initially whipping cables upwards. Because we use fixed-endpoint cables moving at slower speeds, the forces exerted on the rope by gravity and the fixed endpoint prevent us from performing knot ties. 
Kim and Shell~\cite{kim2016using, shen2024diffclip} use a small-scale mobile robot with a cable attached as a tail to manipulate small objects by leveraging inter-object friction to drag or strike the object of interest. While their work aims to tackle dynamic cable control, it relies heavily on physics models for motion primitives that are only applicable to dragging objects via cables, and uses a RRT-based graph search algorithm to address the stochasticity in the system. Other studies have developed physics models to analyze dynamic cable motion. For example, Zimmermann et al.~\cite{zimmermanndynamic} use the finite-element method to model elastic objects for dynamic manipulation of free-end deformable objects, and Gatti-Bonoa and Perkins~\cite{fly_fishing_2004} and Wang and Wereley~\cite{wang2011analysis} analyze fly fishing rod casting dynamics. We focus on fixed-end cable manipulation and do not attempt to model cable physics, and we aim to control cables dynamically via vaulting, knocking, and weaving, and cannot rely on the same dragging motion primitives as in Kim and Shell~\cite{kim2016using}.

We propose an approach that is related to TossingBot~\cite{zeng_tossing_2019} where a robot learns to toss objects to target bins. TossingBot uses RGBD images and employs fully convolutional neural networks~\cite{fcn_2015} to learn dense, per-pixel values of tossing actions, and learns a residual velocity of tossing on top of a physics model in order to jointly train a grasping and tossing policy through trial and error. We take a similar approach in dynamically manipulating by reducing the learning process to learn only the joint angles of the apex of the motion. We do not rely on a physics model of the dynamic manipulation motion, and instead, leverage a quadratic programming-based motion-planning primitive to generate a fast ballistic motion that minimizes the jerk while moving through the learned apex point.

\section{Problem Definition}
Given a manipulator arm holding one end of a cable with the other end fixed (e.g., plugged into a wall), the objective is to compute and execute a single high-speed arm motion from a fixed far-left starting location, to a fixed far-right ending location that moves the cable from an initial configuration to a configuration that matches a task-dependent goal constraint.

Let $s \in \mathcal{S}$ be the configuration of the cable and task-specific objects, including obstacles or targets, where $\mathcal{S}$ is the configuration space.
Since $\mathcal{S}$ is infinite-dimensional, we consider observations of the state, and focus on tasks for which the observations are sufficient to capture progress.  Let $o \in \mathcal{O}$ be the observation, where $\mathcal{O}$ is the observation space and $f_c : \mathcal{S} \rightarrow \mathcal{O}$ map a state to an observation (e.g., via a camera view).
Let $\tau_a \in \mathcal{A}$ be the action the manipulator arm takes, where $\tau_a$ is a complete open-loop trajectory and $\mathcal{A}$ is the action space.
Let $f_\mathcal{S} : \mathcal{S} \times \mathcal{A} \rightarrow \mathcal{S}$ be the stochastic state transition function of the cable.

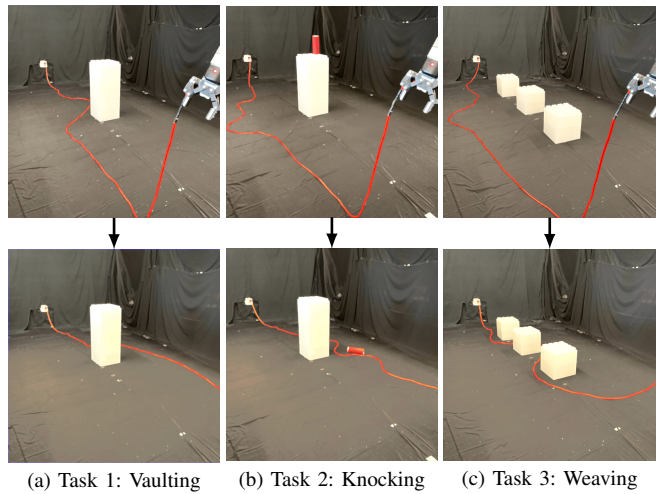
\begin{figure}[t]
\vspace*{5pt}

\center
\begin{tikzpicture}[
every node/.style={outer sep=0,node distance=0pt,inner sep=0},
arr/.style={-latex,line width=1pt}]

\node (vs) {\includegraphics[width=80pt]{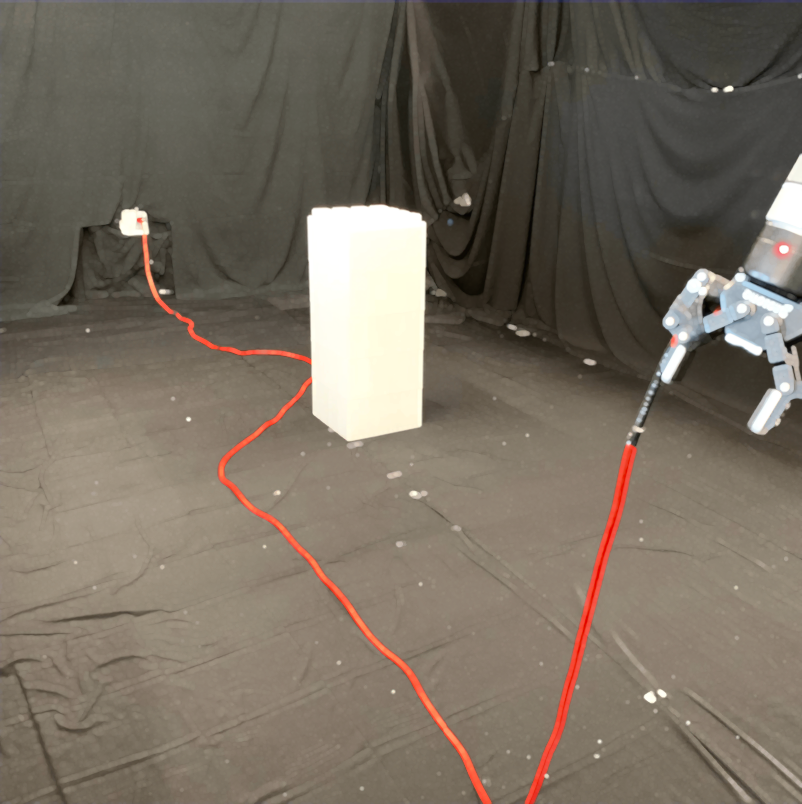}};
\node [right=2.4pt of vs] (ks) {\includegraphics[width=80pt]{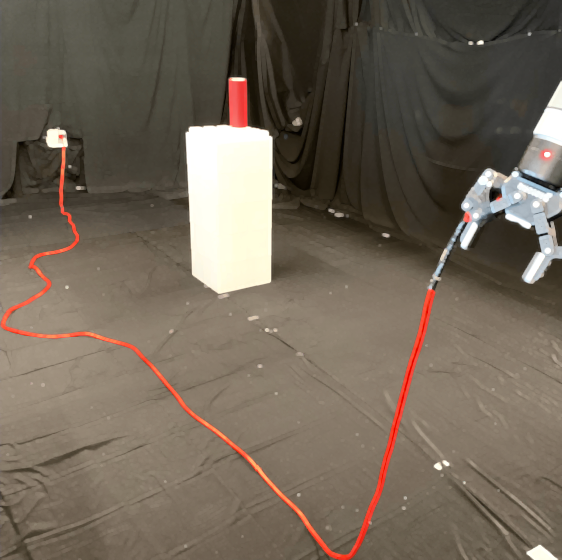}};
\node [right=2.4pt of ks] (ws) {\includegraphics[width=80pt]{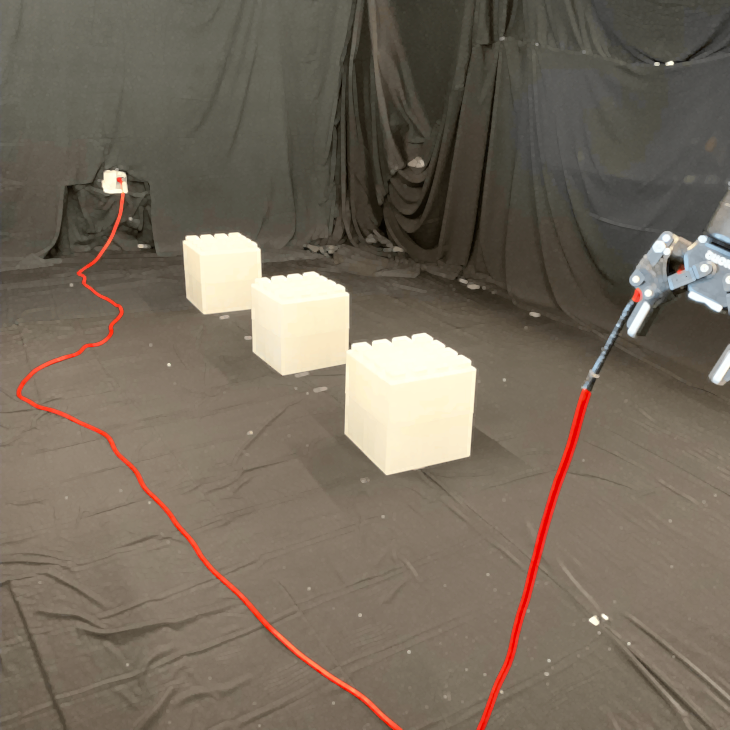}};
\node [below=12pt of vs] (ve) {\includegraphics[width=80pt]{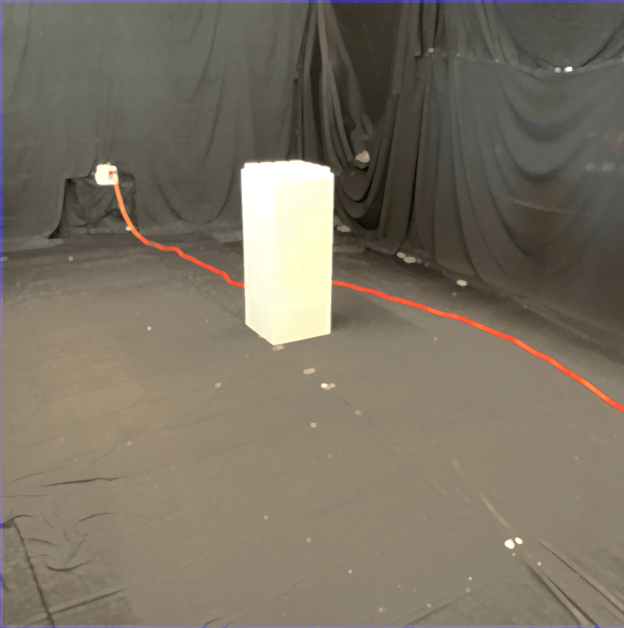}};
\node [below=12pt of ks] (ke) {\includegraphics[width=80pt]{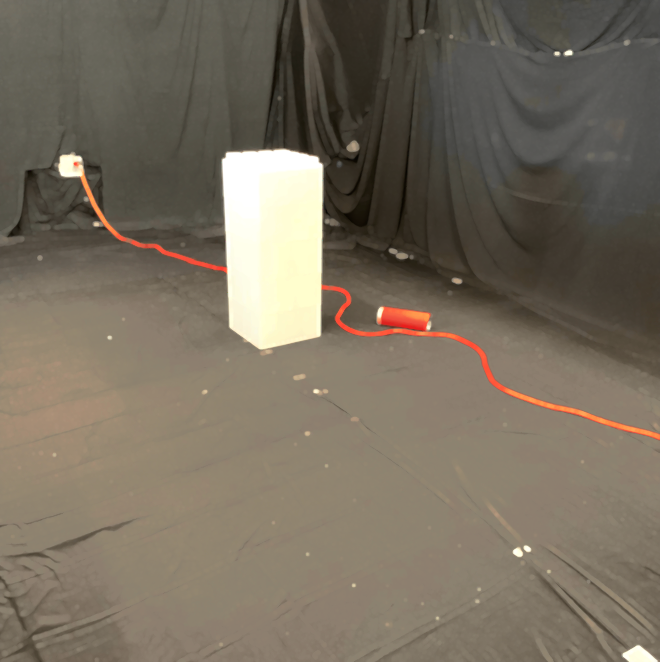}};
\node [below=12pt of ws] (we) {\includegraphics[width=80pt]{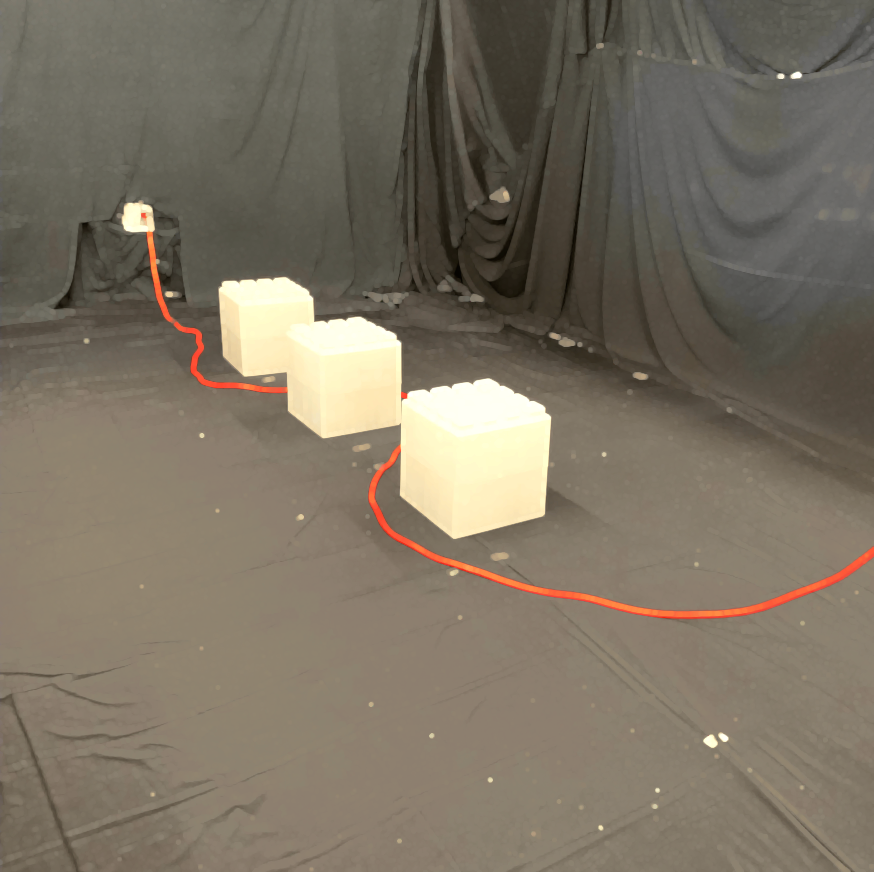}};
\draw [arr] (vs) -- (ve);
\draw [arr] (ks) -- (ke);
\draw [arr] (ws) -- (we);
\node [below=4pt of ve] {\footnotesize (a) Task 1: Vaulting};
\node [below=4pt of ke] {\footnotesize (b) Task 2: Knocking};
\node [below=4pt of we] {\footnotesize (c) Task 3: Weaving};
\end{tikzpicture}
\caption{
\small
Illustration of the 3 tasks: Vaulting, Knocking, and Weaving in real experiments with an orange 18-gauge cable. 
}
\vspace*{-15pt}
\label{fig:three_tasks}
\end{figure}

Formally the objective is: for each task $i \in \{ \mathrm{vaulting},\allowbreak \mathrm{knocking},\allowbreak \mathrm{weaving} \}$, to learn a task-specific policy $\pi_i : \mathcal{O} \rightarrow \mathcal{A}$, that when given the observation $o_0 = f_c(s_0)$ of 
an initial configuration $s_0 \in \mathcal{S}$ and a goal set of observations $\mathcal{O}_\mathrm{goal} \subset \mathcal{O}$, computes an action $\tau_a \in \mathcal{A}$ such that $f_c(f_\mathcal{S}(s_0,\tau_a)) \in \mathcal{O}_\mathrm{goal}$. For each task, we define the goal set in terms of constraints, such as requiring an observation with the cable on the right side of an obstacle (Fig.~\ref{fig:three_tasks}~(a)).


Since the action space $\mathcal{A}$ is also infinite-dimensional, we define a parametric trajectory function $f_{\mathcal{A}_i} : \mathbb{R}^3 \rightarrow \mathcal{A}$, that, when given an apex point, computes complete arm trajectory. While the exact function is a design choice,
we propose that the function should have
the following features: the arm motions are fast enough to induce a motion over the length of the cable, the apex point of the motion can be varied based on an $\mathbb{R}^3$ input,  the other points in the trajectory vary smoothly through the apex, and the motions are kinematically and dynamically feasible. This converts the problem to learning $\hat\pi_i : \mathcal{O} \rightarrow \mathbb{R}^3$, where $\mathcal{O}$ contains position and dimension information, such that for the action $a \in \mathbb{R}^3$ computed by the policy, $f_c(f_\mathcal{S}(s_0, f_{A_i}(a))) \in \mathcal{O}_\mathrm{goal}$.

\begin{figure}[t]
    \vspace*{5pt}
    \centering
    \includegraphics[height=122pt]{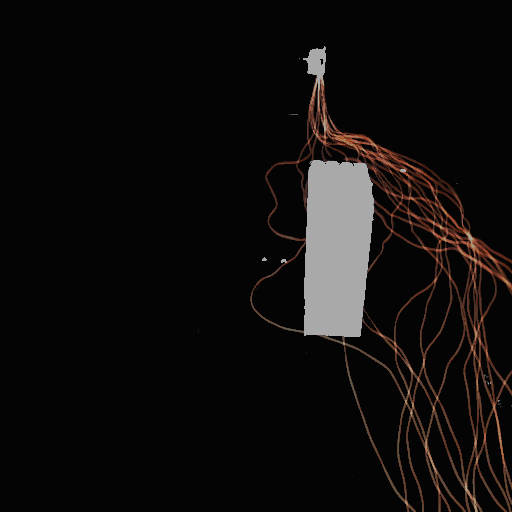}
    \hfill
    \includegraphics[height=122pt]{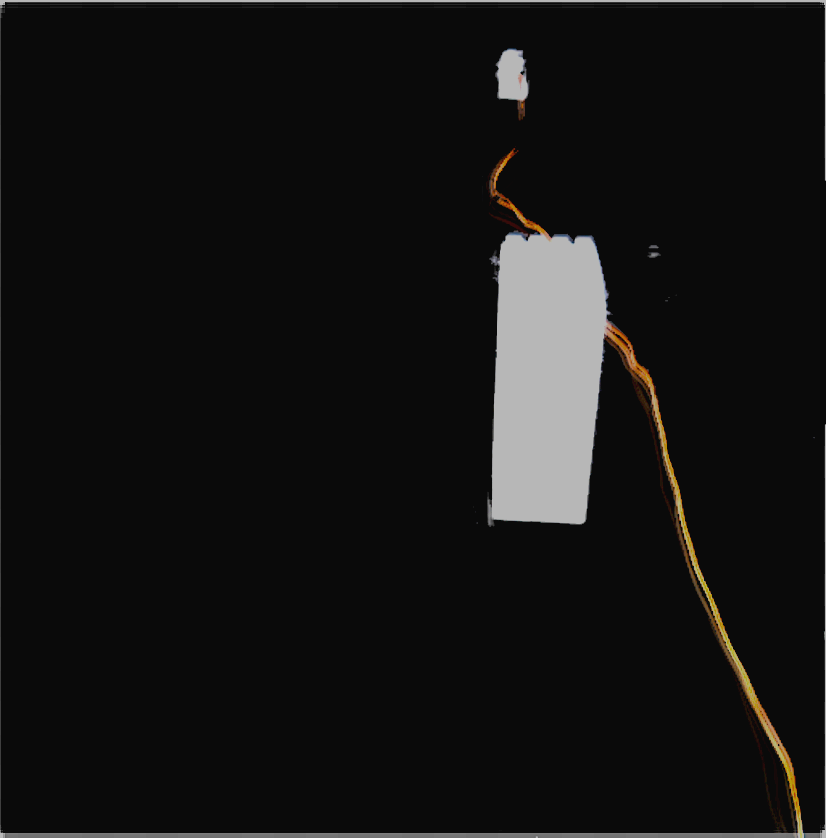}
    \caption{\textbf{Repeatability without (left) and with (right) taut-pull reset.} 
    Both images show an overlay of 20 ending cable configurations after sequentially applying the same left-to-right vaulting trajectory 20 times. \textbf{Left: }without applying resets before the motion, there is high variation in ending configurations.  \textbf{Right: }resetting the cable via a taut-pull before the motion results in nearly identical configurations across 20 left-to-right attempts.}
    \label{fig:repeat}
\end{figure}

\section{Method}
\label{sec:method}

\newcommand{\Tj}[1]{\includegraphics[width=28mm, clip]{figures/#1}}


We propose a deep-learning system that takes as input an image, and produces a sequence of timed joint configurations; when the robot to follows the motion, it results in the dynamic manipulation of a cable that accomplishes the task.

\begin{algorithm}[t!]
\caption{\small Iterative traiNing for Dynamic manipulation (INDy)}
\small{
\begin{algorithmic}[1]
\algsetup{linenosize=\small}

\label{alg:whip}
\REQUIRE Base action of task $i$ $a_{\beta, i}$, a task-trajectory function $f_{\mathcal{A}_i}$, a goal observation set $\mathcal{O}_\mathrm{goal}$, number of data points $N$, the task's time horizon $T$.
\STATE Empty dataset $\mathcal{D} = \{\}$.
\FOR{$t$ $\in \{1 \ldots N\}$}
    \STATE Randomize environment (obstacle size and location)
    \STATE Pull taut to reset to cable's state (updates $s_t$)
    \STATE $o_t \leftarrow f_c(s_t)$ \COMMENT{record initial observation}
    \STATE Set starting action $a_{0:T} = a_{\beta, i}$
    \LOOP
        \STATE $\tau_a \leftarrow f_{\mathcal{A}_i}(a_{0:T})$ \COMMENT{compute trajectory}
        \STATE execute $\tau_a$ on the robot (updates $s_t$)
        \IF{$f_c(s_t)\in\mathcal{O}_\mathrm{goal}$}
            \STATE \textbf{break} \COMMENT{success observed}
        \ENDIF
        \STATE Pull taut to reset to cable's state (updates $s_t$)
        \STATE $a_{0:T} \leftarrow a_{0:T} + \mathrm{(random\;noise)}$
    \ENDLOOP
    \STATE $\mathcal{D} \longleftarrow \mathcal{D} \cup \{(o_t, a_{0:T})\}$ 
\ENDFOR
\STATE Train policy $\pi_i(a|o)$ with $\mathcal{D}$ to minimize Eq.~\ref{eq:loss}.
\RETURN Trained policy for task $i$, $\pi_i(a|o)$
\end{algorithmic}
}
\end{algorithm}
\subsection{Hypothesis}

\paragraph*{Repeatability} A cable with fixed length is attached to the wall and with a gentle pull, the robot can reset the cable to a consistent starting state before each motion.  The same arm trajectory will result in a similar end configuration of the cable (Fig.~\ref{fig:repeat}).

\paragraph*{Parameterized Trajectory Function} We hypothesize that a high-speed minimum-jerk ballistic manipulator-arm motion can solve many left-to-right vaulting problems by varying the midpoint location in $(x,y,z)$ (Fig.~\ref{fig:motion_and_floor}~left).  This reduces the problem to finding which $\mathbb{R}^3$ input to a trajectory function $f_{\mathcal{A}_i}$ solves the task.  This is akin to the $\mathbb{R}^2$-regression TossingBot~\cite{zeng_tossing_2019} uses to throw objects.

\subsection{Deep-Learning Training Algorithm} 
\label{sec:deep_alg}

We propose a self-supervised training algorithm, INDy (Alg.~\ref{alg:whip}), to learn a policy $\pi_i$ for task $i$. The main inputs are:
a base action for the task $a_{\beta,i}$, 
a task-trajectory function $f_{\mathcal{A}_i}$, and
a goal observation set $\mathcal{O}_\mathrm{goal}$.

To train the policy, INDy
first forms a dataset $\mathcal{D}$ of successful observations and actions by having the robot attempt the base action $a_{\beta, i}$.  On failure, it repeatedly tries random perturbations on top of the base action until it successfully completes the task.  Before each attempt, the robot self-resets the cable with a taut-pull, which makes the system repeatable (Fig.~\ref{fig:repeat}). INDy add successful observation-action pairs to $\mathcal{D}$ until the data contains a user-specified number of points $N$. INDy then uses $\mathcal{D}$ to train a convolutional neural network (CNN)~\cite{krizhevsky2012imagenet} policy $\pi_i$ via behavior cloning~\cite{Pomerleau_behavior_cloning,ross2011reduction} to output motion parameters given image observations. We parameterize the policy to produce the mean and covariance of a multivariate Gaussian, and for each data sample $(o_t, a_{0:T}) \sim \mathcal{D}$, optimize:
\begin{equation}\label{eq:loss}
\mathcal{L}_{\mathrm{MSE}} = ||\hat{a}_{0:T} - a_{0:T}||_2^2
\end{equation}
where $\hat{a}_{0:T}\sim \pi_i(a|o_t)$, and optimization is done using the reparameterization trick~\cite{vae}.

\subsection{Min Jerk Trajectory Generation Function}\label{ssec:min-jerk}


INDy uses a trajectory generation function $f_{\mathcal{A}_i} : \mathbb{R}^3 \rightarrow \mathcal{A}$ to reduce the output dimension of the neural network, 
while still being able to achieve a high success rate on tasks.  In initial experiments, we observed that for the trajectory to be dynamically feasible, the trajectory generation must also limit and minimize jerk.  Otherwise, acceleration changes would often be too fast for the manipulator arm to follow while burdened by the inertia of the cable.

We propose a function based on a quadratic program (QP) that has these features and empirically achieves high success rates in the 3 tasks.  
We first discretize a trajectory to $H+1$ time steps, each with $h$ seconds apart.  At each time step $t \in \irange 0H$, the discretized trajectory has a configuration $\vec{q}_t$, and a configuration-space velocity $\vec{v}_t$ and acceleration $\vec{a}_t$.  With constraints in the QP, we enforce that the trajectory is kinematically and dynamically feasible, starts and ends at 2 fixed arm configurations, and reaches an apex configurations defined by the output of the learned policy.  The objective of the QP minimizes jerk.  To make the trajectory as fast as possible, in a manner similar to GOMP~\cite{GOMP}, we repeatedly reduce $H$ by 1 until the QP is infeasible, and use the shortest feasible trajectory.  The QP takes the following form:
\begin{small}
\begin{align*}
    \argmin_{(\vec{q},\vec{v},\vec{a})_{\irange 0H}} \; & \frac 1{2h} \sum_{t=0}^{H-1} (\vec{a}_{t+1} - \vec{a}_{t})^{\intercal} Q (\vec{a}_{t+1} - \vec{a}_{t}) \\
    \text{s.t.} \; & \vec{q}_0 \text{ is the start configuration} \\
    & \vec{q}_{\lfloor H/2 \rfloor} \text{ is the apex configuration} \\
    & \vec{q}_H \text{ is at the end configuration} \\
    & \vec{v}_0 = \vec{v}_{H} = \vec{0} \\
    & \vec{q}_{t+1} = \vec{q}_t + h \vec{v}_t + \tfrac 12 h^2 \vec{a}_t \quad \forall t \in \irange 0{H-1} \\
    & \vec{v}_{t+1} = \vec{v}_t + h \vec{a}_t \quad \forall t \in \irange 0{H-1} \\
    & \vec{q}_\mathrm{min} \le \vec{q}_t \le \vec{q}_\mathrm{max} \quad \forall t \in \irange 1{H-1} \\
    & \vec{v}_\mathrm{min} \le \vec{v}_t \le \vec{v}_\mathrm{max} \quad \forall t \in \irange 1{H-1} \\
    & \vec{a}_\mathrm{min} \le \vec{a}_t \le \vec{a}_\mathrm{max} \quad \forall t \in \irange 0{H-1} \\
    & h \vec{j}_\mathrm{min} \le \vec{a}_{t+1} - \vec{a}_t \le h\vec{j}_\mathrm{max} \quad \forall t \in \irange 0{H-1}.
\end{align*}
\vspace*{-15pt}
\end{small}

The $Q$ matrix is a diagonal matrix that adjusts the relative weight of each joint.  The QP constraints, in order, fix the start, midpoint, and end configurations ($\vec{q}_0$, $\vec{q}_{\lfloor H/2 \rfloor}$, $\vec{q}_{H}$); fix the start and end velocity ($\vec{v}_0$, $\vec{v}_{H}$) to zero; enforce configuration and velocity dynamics between time steps ($\vec{q}_{t+1}$, $\vec{v}_{t+1}$); keep the joints within the specified limits of configuration ($\vec{q}_\mathrm{min}$, $\vec{q}_\mathrm{max}$), velocity ($\vec{v}_\mathrm{min}$, $\vec{v}_\mathrm{max}$), acceleration ($\vec{a}_\mathrm{min}$, $\vec{a}_\mathrm{max}$), and jerk ($\vec{j}_\mathrm{min}$, $\vec{j}_\mathrm{max}$).
Unlike GOMP, we do not consider obstacle avoidance in the trajectory generation, and instead ensure that there are no obstacles in the path of the arm to avoid.  In practice, we fix $h$ to be an integer multiple of the robot's control frequency, and the initial value of $H$ to be sufficiently large such that there is always a solution.

The input to the trajectory function constrains the midpoint configuration of the trajectory.  As most trajectories lift then lower the end effector through the midpoint, the input is often the apex of the motion.  With the UR5 robot, we observe that varying the first 3 joints (base, shoulder, elbow) moves the end-effector's location in $(x,y,z)$.  The base joint rotates left and right, while the shoulder and elbow joints lift and extend.  We thus use these 3 joints to define the midpoint, while having the remaining joint configurations depend linearly on the first three.
We make this function a task-specific function by constraining the start and end configurations to depend on the task.  For example, for the left-to-right vaulting task, we set the start configuration to be down and left, and the end configuration to be down and right.

\section{Simulatior}
\label{sec:sim2real}

We use simulation to experiment with training algorithms, and to find a feasible base action $a_\beta$ to bootstrap the data collection process in the real world.  Using simulation enables fast prototyping to efficiently assess qualitative and quantitative behavior of multiple base actions, but can suffer from a sim-to-real gap, motivating research teams to collect data from running robots in the real world~\cite{gupta2018,folding_fabric_fcn_2020}. Similarly, we collect data by running a physical UR5 robot, but use simulation as a means to obtain a functional action that can be improved upon when collecting data.

To model cables in simulation, we use a Featherstone~\cite{featherstone1984robot} rigid-body simulator from BulletPhysics via Blender~\cite{blender}. Cables consist of a series of small capsule links connected by spring and torsional forces. 
We tune the spring and torsional coefficients by decreasing bending resistance and stopping before the cable noticeably behaves like individual capsules rather than a cable, and increasing twist resistance until excessive torsion causes the capsules to separate.

\begin{figure}
    \vspace*{5pt}
    \centering
    \begin{tabular}{@{}c@{\hspace{2.4pt}}c@{\hspace{2.4pt}}c@{}}
        \includegraphics[width=80pt]{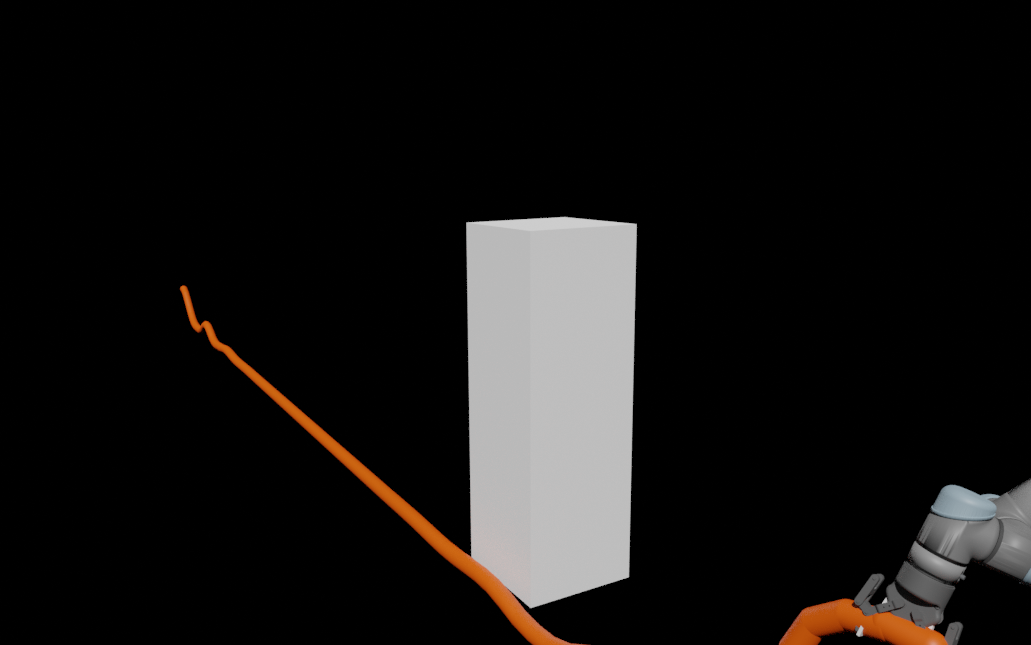} & \includegraphics[width=80pt]{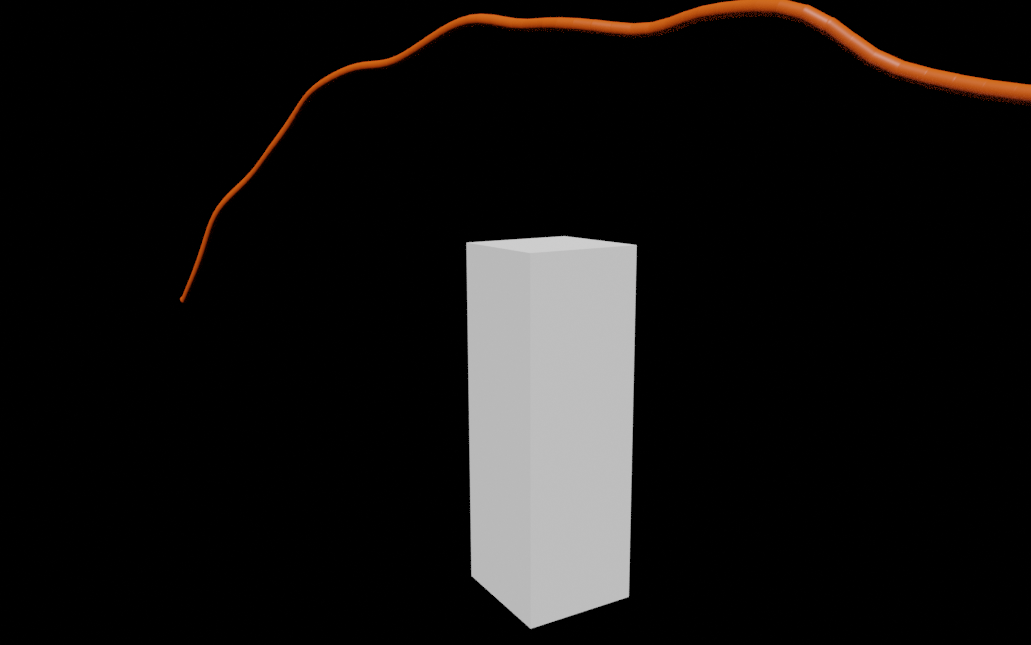} & \includegraphics[width=80pt]{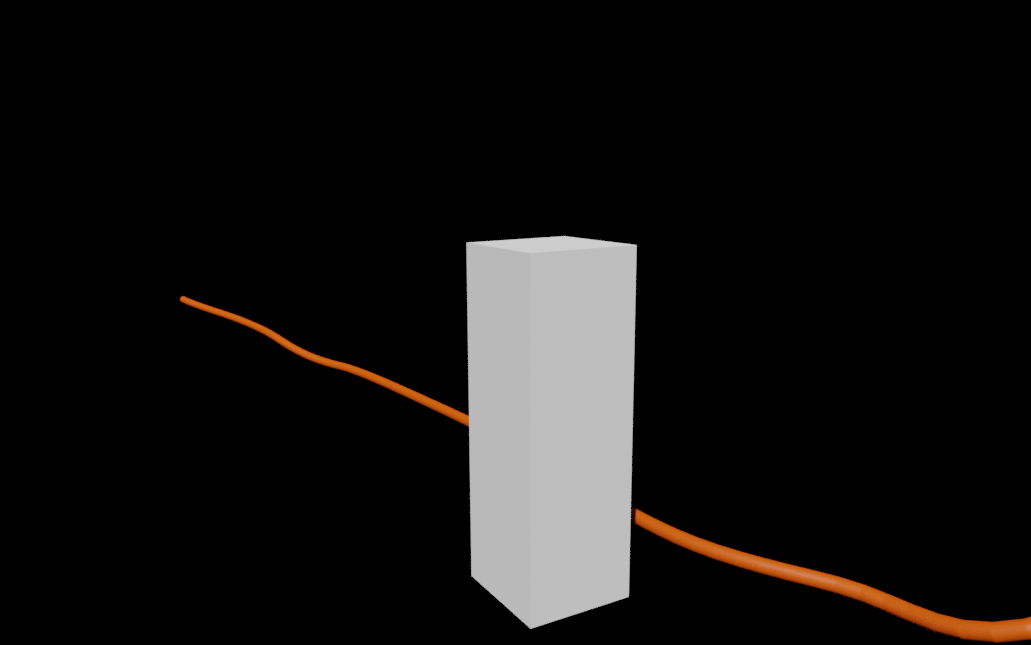} \\
        \includegraphics[width=80pt]{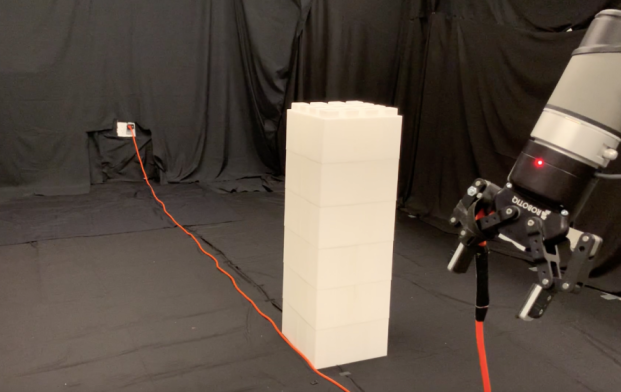} & \includegraphics[width=80pt]{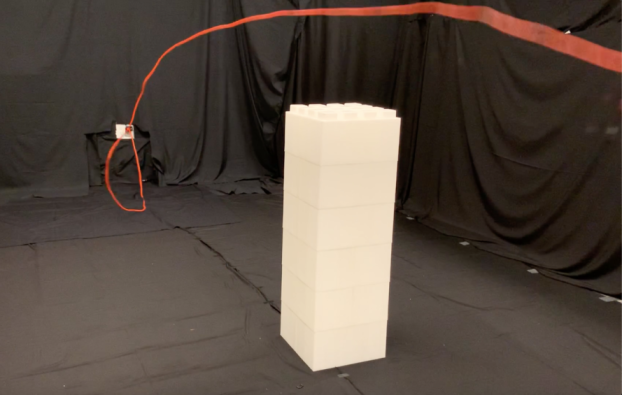} & \includegraphics[width=80pt]{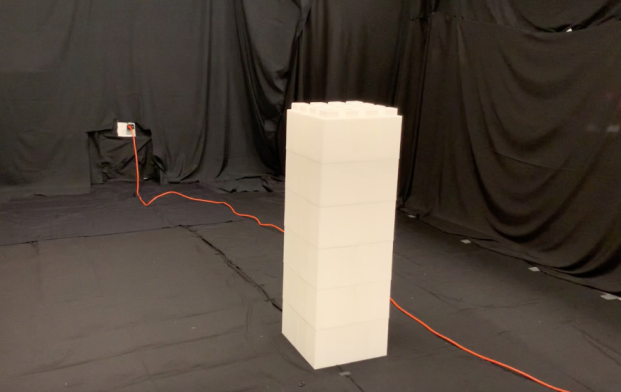}\\
        \small{Start configuration} & \small{Apex configuration} & \small{End configuration}
    \end{tabular}
    \caption{Cable trajectory for data collection in \textbf{simulation (top)} vs. cable trajectory in \textbf{real (bottom)} for the vaulting task after applying the same apex point configuration.}
    \label{fig:sim_to_real}
    \vspace*{-10pt}
\end{figure}

When finding a feasible $a_\beta$ to transfer to real, we only simulate vaulting. The obstacle height, radius, and location are randomized for each trial, along with the initial cable state. We use 10 different obstacle and initial cable settings in simulation, and select the input apex point with highest success rate out of 60 random points. Then, we take this apex point directly to physical experiments as the base action $a_\beta$ to collect data, as the physical UR5 exhibits similar behavior to that in simulation after applying the predefined configurations from simulation, as seen in Fig.~\ref{fig:sim_to_real}.




\section{Physical Experiments}

\begin{figure}[t]
    \vspace*{5pt}
    \centering
      \includegraphics[height=110pt]{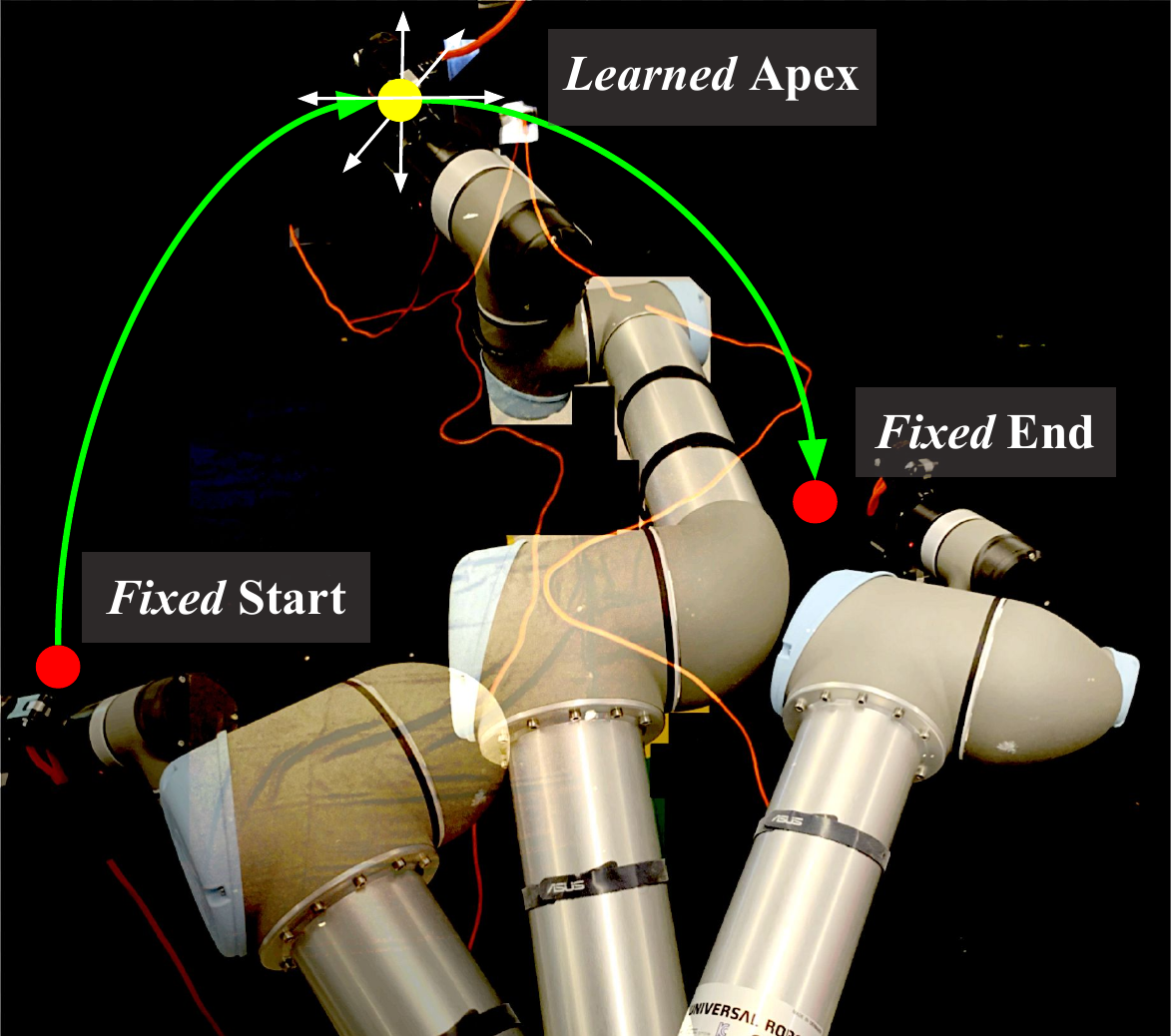}
      \hfill
      \includegraphics[height=110pt]{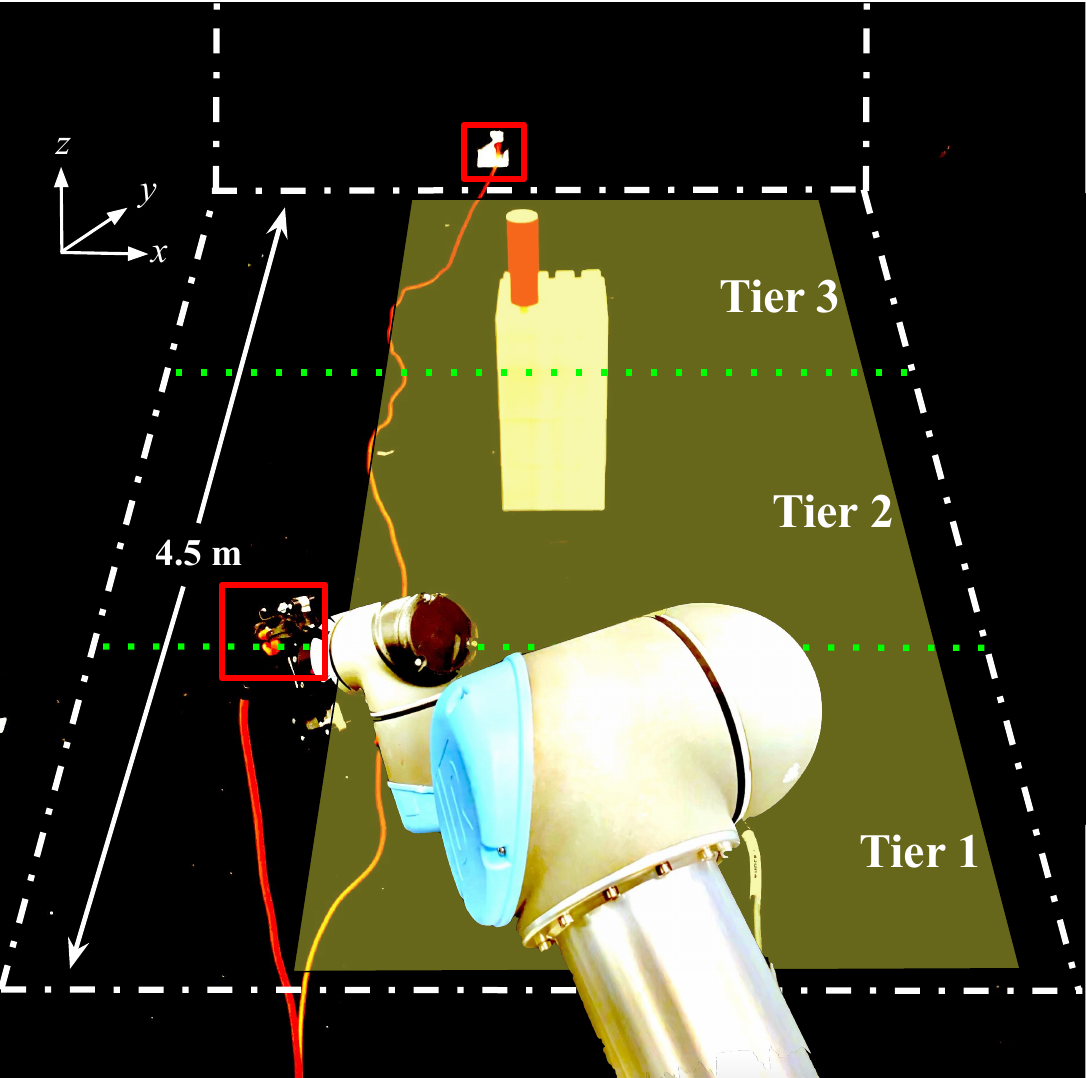}
    \caption{\textbf{Left: }Using three points in a curve to control the trajectory of the cable. The start and end points are fixed using the arm configurations. We change the trajectory by varying the apex configuration. \textbf{Right: }Design of physical experiments. We randomize the white Lego bricks’ location within the yellow shaded area. For vaulting and knocking, the width of the yellow area is 1.5~m, and 1.0~m for weaving. The yellow area is divided into three difficulty tiers.}
    \label{fig:motion_and_floor}
    \vspace*{-10pt}
\end{figure}

We use a physical UR5 robot grasping to dynamically manipulate a cable attached to a wall 4.5\,m away, on the three tasks, with 5 different cable types.
The system obtains observations from a Logitech C270 720p webcam placed 0.75\,m above the robot arm base.  It scales and crops the image to a 512x512x3 segmented RGB image that feeds into a ResNet-34~\cite{resnets_2016} deep neural network implemented with PyTorch~\cite{pytorch_neurips}.
We initialize weights using He initialization~\cite{he2015delving}.

In physical experiments of the 3 tasks (see Fig.~\ref{fig:motion_and_floor} for layout),
we use large, white plastic bricks as the obstacle(s), which can vary in size and location. The obstacles we use for experiments are 0.15\,m to 0.75\,m in width and 0.3\,m to 1.5\,m in height.
To generate training data, we randomize the location and size of the obstacles in each trial.
We test the system with 5 cables listed in Table~\ref{tab:cable_types}.

\begin{table}[t]
    \vspace*{4pt}
    \centering
    \caption{Physical cables used in experiments and the number of successes in the three tasks (Vaulting, Knocking, Weaving) using policies trained with the 18 AWG orange cable. For each cable, we evaluate with 20 trials for each task.}
    \begin{tabular}{@{}l@{\quad}r@{\quad}r@{\quad}r@{\quad}r@{\quad}r@{}}\toprule
        
         Type & kg & m & Vaulting & Knocking & Weaving \\
        \midrule
         18 AWG orange cable  & 0.65 & 5.5 & 16&13&12\\
         16 AWG white cable   & 0.70 & 5.2 &16&12&12\\
         16 AWG orange cable  & 0.90 & 6.2 &4&2&2\\
         22 AWG blue network cable  & 0.48 & 5.5 &14&9&12\\
         12 AWG blue jump rope   & 0.45 & 5.1 &15&10&8\\
         \bottomrule
    \end{tabular}
    \vspace*{-16pt}
    \label{tab:cable_types}
\end{table}


We define \emph{difficulty tiers} based on the distance of the obstacle to the base of the robot.
%
\begin{table}[h]
\centering
\vspace{-6pt}
\begin{tabular}{@{}ccc@{}}
    \textbf{Tier 1} & 
    \textbf{Tier 2} & 
    \textbf{Tier 3} \\
    ${\pmb{\leq}}$ 1.5\,m &
    1.5 to 3\,m & 
    ${\pmb{\geq} }$ 3\,m
\end{tabular}
\vspace{-12pt}
\end{table}
%
%
%

We evaluate INDy along with two baselines:
\begin{description}[align=left,leftmargin=0pt]
    \item[Baseline with Fixed Apex.] In the three tasks, for each configuration, a human annotator sets one set of apex joint angles that can accomplish the task. The apex joint angles are fixed with respect to the obstacle itself, and they remain the same across different locations of the same obstacle.
    \item[Baseline with Varying Apex.] In vaulting and weaving, for each configuration (including obstacle size and location), a human annotator sets the apex joint angles such that the robot arm's end effector is 15\,cm above the center of mass (COM) of the obstacle, and aligned horizontally with the COM of the obstacle. For knocking, a human annotator makes the robot arm end 10\,cm above the COM of the target object, and aligned horizontally with the COM.
\end{description}

We find the baselines from simulation, where we observe that the fixed and varying apex motions could induce enough velocity to make the cable vault the obstacle.
\section{Results}

\begin{figure}[t]
    \vspace*{15pt}
    \centering
    \includegraphics[width=120pt]{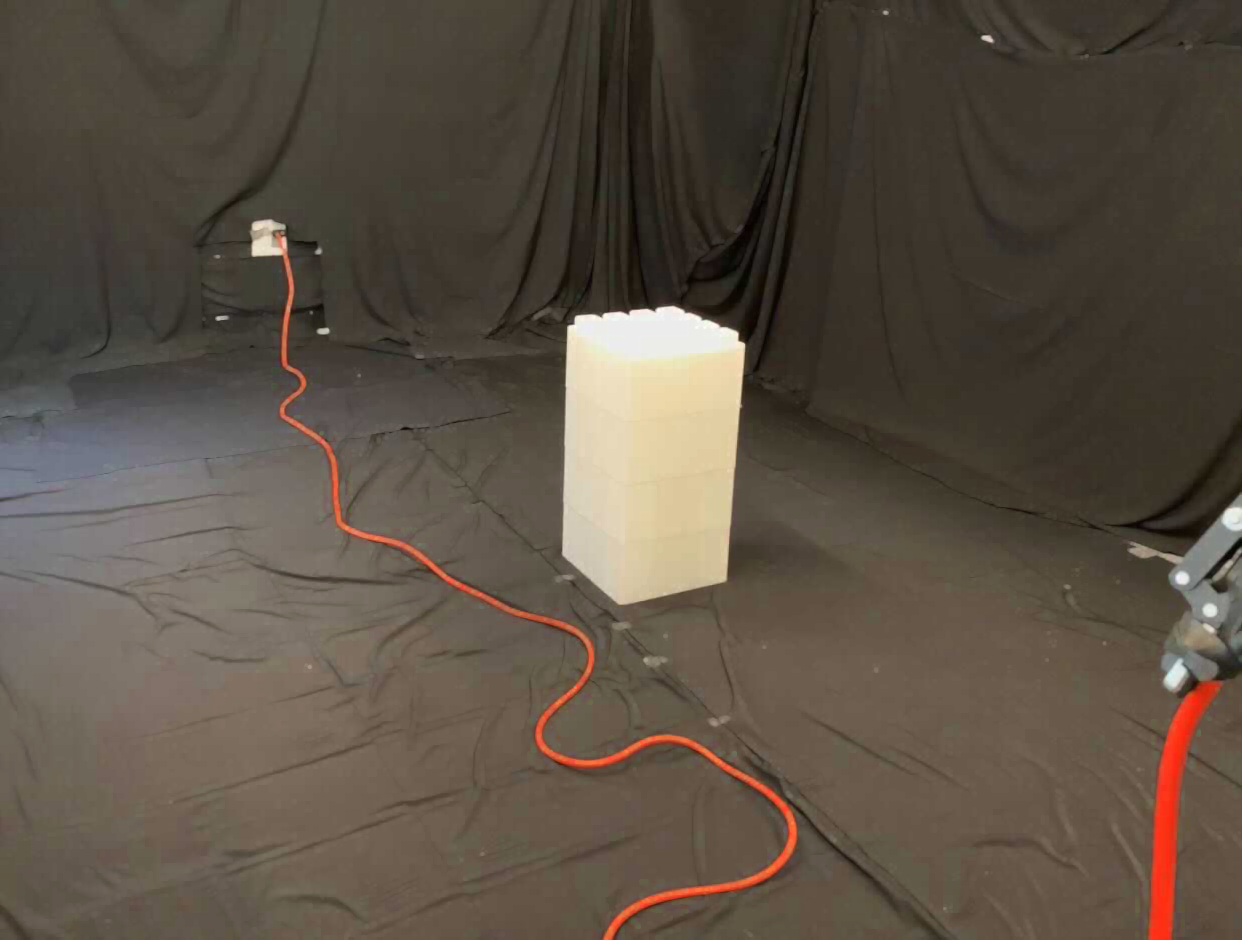}\hfill
    \includegraphics[width=120pt]{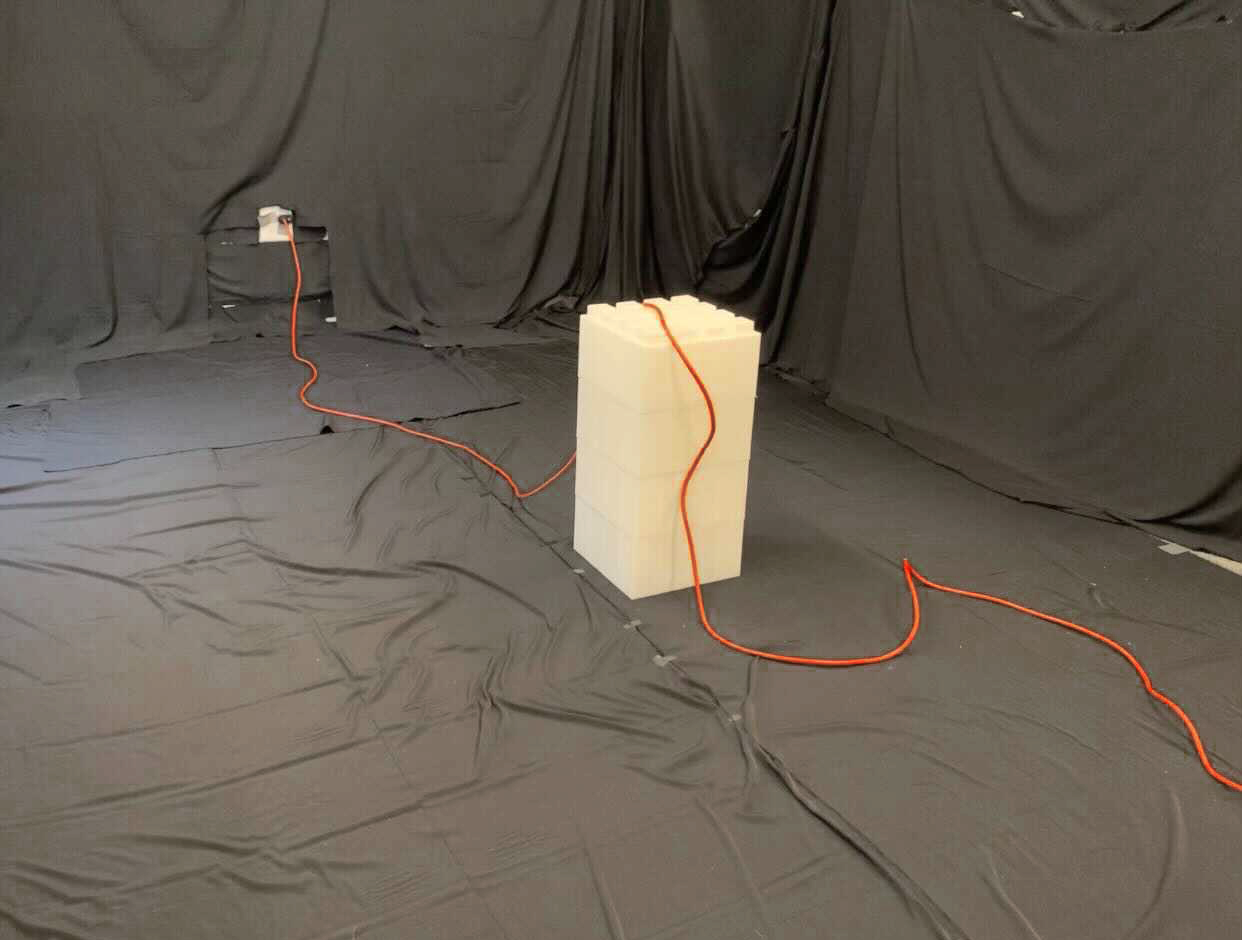}
    \caption{\textbf{Failure mode}. The 16-gauge 6.2~m orange cable is too long for the vaulting motion.  \textbf{Left:} before the motion there is too much slack. \textbf{Right:} the vaulting motion that works for shorter cables does not clear the obstacle.} 
    
    \label{fig:failure}
\end{figure}


We measure the correlation between simulated and real setups, benchmark the three dynamic cable tasks, and investigate generalization to different cables. 
The project website supplements these results with videos.

\subsection{Comparisons between Simulation and Reality}

To observe correlations between cable behavior in simulation and real, we evaluate vaulting on the same set of apex point and obstacle settings in both setups. Each trial consists of a random obstacle location from all 3 difficulty tiers and a different apex point. In simulation, we use a cube mesh of the same dimension as the plastic bricks obstacle in real, and a cable of the same length and mass as the 18 AWG orange cable. Table~\ref{table:sim_to_real} reports the frequency of success and failure cases across simulation and real. While the cable trajectory in simulation tends to match the trajectory of real for successes, 50\,\% of all failed cases in real succeed in simulation under the same setting. The discrepancies in cable behavior between simulation and real are most apparent for Tier 3 scenarios, where the height achieved by the cable at the opposite end from the robot differ in the two environments.  Because the cable properties cannot be perfectly tuned to match the physical properties in real, these experiments motivate the decision to use real experiments to collect data.


 
 \begin{table}
    \vspace*{4pt}
\centering
\caption{Number of successes and failures for vaulting in simulation and physical environments across 15 trials, evaluating a different apex point on a different obstacle position for each trial. Listed are the number of cases that succeed in both sim and real, fail in sim but succeed in real, succeed in sim but fail in real, and fail in both sim and real. The correlation between success rates in the two environments is surprisingly high.}
 \begin{tabular}{ 
l c c}
 \toprule
   & \#Success (Simulation) & \#Failure (Simulation) \\
  \midrule
  \#Success (Physical)&7&2\\
  \#Failure (Physical)&3&3\\
 \bottomrule
\end{tabular}
\vspace*{-5pt}

\label{table:sim_to_real}
\end{table}
\subsection{Results for Vaulting Task}
For vaulting, we perform 180 physical trials from manipulating the orange 5.5\,m 18-gauge power extension cable, with 60 trials for each of the 3 difficulty tiers; this process took 8 hours. In each tier, we shift the obstacle vertically within the bounds of the difficulty tier and horizontally within a range of 1.5\,m. We use plastic bricks to construct 6 combinations to build 6 objects of different sizes, and for each obstacle, we vary its location 10 times. Table~\ref{table:res_task1} shows results of the trained policy on vaulting evaluated with 20 trials per tier.

\begin{table}[t]
\vspace*{0pt}
\centering
\caption{Physical Vaulting Results: Number of successes of the learned policy for vaulting evaluated against two baseline methods across 20 trials per difficulty tier using the 18 AWG orange cable.}
 \begin{tabular}{ 
l c c c}
 \toprule
  Method & Tier 1 & Tier 2 & Tier 3\\
  \midrule
  Baseline with Fixed Apex&16&13&2\\
  Baseline with Varying Apex&19&14&7\\
  Learned Apex &\textbf{20}&\textbf{18}&\textbf{11}\\
 \bottomrule
\end{tabular}

\label{table:res_task1}
\end{table}


\subsection{Results for Knocking Task}

For knocking, we perform 180 physical trials from manipulating the orange 5.5-m 18-gauge power extension cable, with 60 trials for each of 3 difficulty tiers; this process took 9 hours. In each difficulty tier, we use the plastic bricks as the base upon which we place the target object to be knocked, and we shift the plastic bricks base vertically within the bounds of the difficulty tier and horizontally within a range of 1.5\,m. We train the policy using four different target objects: a cylinder, a tennis ball, a cup, and a rectangular box. We use plastic bricks to construct 3 combinations to build 3 bases of different sizes. For each base, we vary its locations 5 times, and for each location of the base, we vary the target on top 4 times. 
Table~\ref{table:res_task2} shows the results for the knocking trained policy evaluated with 20 trials for each difficulty tier.
\begin{table}
\vspace*{4pt}
\centering
\caption{Physical Knocking Results: Number of successes of the learned policy for knocking evaluated against two baseline methods across 20 trials per difficulty tier using the 18 AWG orange cable.}
 \begin{tabular}{ 
l c c c}
 \toprule
  Method & Tier 1 & Tier 2 & Tier 3\\
  \midrule
  Baseline with Fixed Apex &12&6&4\\
  Baseline with Varying Apex &14&14&6\\

  Learned Apex&\textbf{15}&\textbf{14}&\textbf{10}\\
 \bottomrule
\end{tabular}
\label{table:res_task2}
\end{table}

\subsection{Results for Weaving Task}

We perform 100 physical trials from manipulating the orange 5.5-m 18-gauge power extension cable, which took 7.5 hours. We make the 3 obstacles identical in size. In our training data for this task, we vary the space between each object by 15 to 50\,cm and we shift the obstacles horizontally within a range of 1\,m.
Table~\ref{table:res_task3} shows the result of the weaving trained policy evaluated with 20 trials. 

\subsection{Generalization to Different Cables}
While the policies for the three tasks are trained using the 18-gauge orange cable, we experiment with different types of cables. The results suggest that the policy trained on the 18-gauge orange cable can transfer well to other cables of similar lengths. We observe that the same reset action can bring these cables to nearly identical starting configurations in observation space, making the policy agnostic to cable type. We use the learned policy from the 18-gauge orange cable without retraining or finetuning to run the same experiments (same object sizes and locations for each task) for the 3 tasks with different cables and show results in Table~\ref{tab:cable_types}. 
\begin{table}
\vspace*{0pt}
\centering
\caption{Physical Weaving Results: Number of successes of the learned policy for weaving evaluated against two baseline methods across 20 trials using the 18 AWG orange cable.}
 \begin{tabular}{ 
l c c c}
 \toprule
  Method & \#Success\\
  \midrule
  Baseline with Fixed Apex &3\\
  Baseline with Varying Apex &3\\

  Learned Apex&\textbf{12}\\
 \bottomrule
\vspace*{-10pt}
\end{tabular}
\label{table:res_task3}

\end{table}
\vspace*{-8pt}
\subsection{Limitations and Failure Modes}

While the framework can be robust to object locations and shapes, it has limitations. First, the policies are learned in a highly controlled environment. 
Many failure cases in vaulting and knocking are from 
placements outside the training area. For example, in Tier 3 vaulting, 7 of 9 failure cases are due to the horizontal shift of the obstacle exceeding the 1.5\,m bound. In weaving, the horizontal shift of the obstacles is bounded into a smaller 1.0\,m range.
%
Second, learned policies have difficulty generalizing to cables of different lengths and masses. We conjecture that with different lengths, the amount of slack in the cables should also be different, thus decreasing the repeatability of applying the same action. With the 6.2\,m cable, Tier 2 and 3 vaulting and knocking always fail in that the segment of the cable near the wall end can never be accelerated up due to the fact that the slack near the wall end cannot be eliminated by the resetting action.
To show this failure mode with a longer cable, we run the policies for vaulting, knocking, and weaving trained with the 5.5-m 18-gauge cable on the 6.2-m 16-gauge cable, and we achieve 20\,\% success rate for vaulting, 10\,\% success rate for knocking, and 10\,\% success rate for weaving. Fig.~\ref{fig:failure} shows this failure case. Additionally, Table~\ref{tab:cable_types} suggests that cables that are too light are likely to yield lower success rates, which may be because they often travel slower during descent, limiting the distance the cable traverses. Finally, the data generation part in Alg.~\ref{alg:whip} is inefficient since when the obstacle size and location change, the algorithm needs to search for successful apex point again, so data generation in real is unable to record a wider variety of obstacle placements in an efficient manner.


\section{Conclusion}

This paper explores learning three dynamic cable manipulation tasks:  vaulting, knocking, and weaving. Experimental results suggest that the proposed procedure outperforms two baseline methods.

In future work, we will generalize the learned policy to different cable lengths and masses, and more complex obstacles and target objects. We will apply the parameterized trajectory method to other dynamic cable manipulation tasks without fixed-endpoints of the cable, inspired by the adventures of Dr. Indiana Jones.

\section{Acknowledgments}

\footnotesize
This research was performed at the AUTOLAB at UC Berkeley in
affiliation with the Berkeley AI Research (BAIR) Lab, the CITRIS ``People and Robots'' (CPAR) Initiative, and the Real-Time Intelligent Secure Execution (RISE) Lab. The authors were supported in part by donations from Toyota Research Institute and by equipment grants from PhotoNeo, NVidia, and Intuitive Surgical. Daniel Seita is supported by a Graduate Fellowship for STEM Diversity (GFSD). We thank our colleagues who provided helpful feedback, code, and suggestions, in particular Francesco Borrelli, Adam Lau, Priya Sundaresan, and Jennifer Grannen.

\bibliographystyle{IEEEtran}
\bibliography{IEEEabrv,references}

\clearpage

\end{document}